\documentstyle[11pt]{article}

\newenvironment{reverseindent}%
{\begin{list}{}{\setlength{\labelsep}{0in}
	        \setlength{\labelwidth}{0in}
	        \setlength{\itemindent}{-\leftmargin}}}%
{\end{list}}

\begin{document}

\begin{center}
{\Large\bf Training and Scaling Preference Functions \\
for Disambiguation}\\[5mm]
{\large Hiyan Alshawi}\\[2mm]
AT\&T Bell Laboratories\\
600 Mountain Ave.,
Murray Hill, New Jersey 07974\\[1mm]
{\tt hiyan@research.att.com}\\[3mm]
{\large David Carter}\\[2mm]
SRI International\\
Cambridge Computer Science Research Centre\\
23 Millers Yard, Cambridge CB2 1RQ, England\\[1mm]
{\tt dmc@cam.sri.com}\\[3mm]
Archive number: cmp-lg/9408013\\[2mm]
{\bf Topics:} preference, collocations, disambiguation\\[3mm]
{\bf ABSTRACT}
\end{center}
\noindent
{\small We present an automatic method for weighting the contributions
of preference functions used in disambiguation.  Initial scaling factors
are derived as the solution to a least-squares minimization problem,
and improvements are then made by hill-climbing.  The method is
applied to disambiguating sentences in the ATIS (Air Travel
Information System) corpus, and the performance of the resulting
scaling factors is compared with hand-tuned factors. We then focus on
one class of preference function, those based on semantic lexical
collocations. Experimental results are presented showing that such
functions vary considerably in selecting correct analyses. In particular
we define a function that performs significantly better than ones based
on mutual information and likelihood ratios of lexical associations.}

\section{Introduction}
\label{intro}

The importance of good preference functions for ranking competing
analyses produced by language processing systems grows as the coverage
of these systems improves. Increasing coverage usually also increases
the number of analyses for sentences previously covered, bringing the
danger of lower accuracy for these sentences. Large scale rule based
analysis systems have
therefore tended to employ a collection of functions to produce a score
for sorting analyses in a preference order.  In this paper we address
two issues relating to the application of preference functions.

\subsection*{Combining Multiple Preference Functions}

The first problem we address is that of combining different functions,
each of which is supposed to offer some contribution to selecting the
best among a set of analyses of a sentence. Although multiple functions
have been used in other systems (for example McCord 1990, Hobbs and
Bear 1990), little is typically said about how the functions are
combined to produce the overall score for an analysis, the weights
presumably being determined by intuition or trial and error.  McCord
(1993) gives very specific information about the weights he uses
to combine preference functions, though these weights are chosen
by hand.  Selecting weights by hand, however, is a task for experts,
which needs to be redone every time the system is applied to a new
domain or corpus. Furthermore, there is no guarantee that the
selected weights will achieve optimal or even near-optimal performance.

The speech processing community, on the other hand, has a longer
history of using numerical evaluation functions, and speech
researchers have used schemes for scoring recognition hypotheses
that are similar to the one proposed here for disambiguation.
For example, Ostendorf et al. (1991) improve recognition performance
by using a linear combination of several scoring functions.
In their work the weights for the linear combination are chosen
to optimize a generalized mean of the rank of the correct word
sequence.

In our case, the problem is formulated as follows. Each preference
function is defined as a numerical (possibly real-valued) function on
representations corresponding to the sentence analyses. A weighted sum
of these functions is then used as the overall measure to rank the
possible analyses of a particular sentence. We refer to the
coefficients, or weights, used in this linear combination as the
``scaling factors'' for the functions.  We determine these scaling
factors automatically in order both to avoid the need for expert
hand-tuning and to achieve performance that is at least locally
optimal.  We start with the solution to minimizing a squared-error
cost function, a well known technique applied to many optimisation
and classification problems. This solution is then enhanced by
application of a hill-climbing technique.

\subsection*{Word Sense Collocation Functions}

Until recently, the choice of the various functions used in rule
based systems was made mainly according to anecdotal information
about the effectiveness of, for example, various attachment preference
strategies. There is now more empirical work comparing such
functions, particularly in the case of functions based on statistical
information about lexical or semantic collocations. Lexical
collocation functions, especially those determined statistically,
have recently attracted considerable attention
in computational linguistics (Calzolari and Bindi 1990, Church and
Hanks 1990, Sekine et al. 1992, and
Hindle and Rooth 1993) mainly, though not exclusively,
for use in disambiguation.
These functions are typically derived by observing the occurrences of
tuples (usually pairs or triples) that summarise relationships present
in an analysis of a text, or their surface occurrences. For example,
Hindle and Rooth (1993) and Resnik and Hearst (1993) give
experimental results on the effectiveness of functions based
on lexical associations, or lexical-class associations, at
selecting appropriate prepositional phrase attachments.

We have experimented with a variety of specific functions which
make use of collocations between word senses. The results we present
show that these functions vary considerably in disambiguation accuracy,
but that the best collocation functions are more effective than a function
based on simple estimates of syntactic rule probabilities.  In
particular, the best collocation function performs significantly better
than a related function that defines collocation strength in terms of
mutual information, reducing the error rate in a disambiguation task
from approximately 30\% to approximately 10\%.

We start by describing our experimental context and training
data in section \ref{setup}. Then we address the issue of
selecting scaling factors by presenting our optimisation procedure in
section \ref{scaling} and a comparison with manual
scaling in section \ref{analysis}. Finally we take a close look
at a set of semantic collocation functions, defining them in
in section \ref{semcoll} and comparing their effectiveness
at disambiguation in section \ref{collcompare}.

\section{The Experimental Setup}
\label{setup}

\subsection*{Disambiguation Task}

All the experiments we describe here were done with the Core Language
Engine (CLE) a primarily rule based natural language processing system
(Alshawi 1992). More specifically, the work on optimising preference
factors and semantic collocations was done as part of a project on
spoken language translation in which the CLE was used for analysis and
generation of both English and Swedish (Agn\"{a}s {\it et al}, 1993).
The work presented here is all concerned with the English analysis
side, though we see no reason why its conclusions should not be
applicable to Swedish or other natural languages.

In our experiments we made use of the
Air Travel Information System (ATIS) corpus of transcribed speech
sentences. This application was chosen because
the proposed method for automatic derivation of scaling factors
requires a corpus of sentences that are representative of the
sublanguage, together with some independent measure of the correctness
or plausibility of analyses of these sentences,
and we had access to a hand-parsed sub-collection of the ATIS
corpus built as part of the Penn Treebank project (Marcus,
Santorini and Marcinkiewicz 1993).
Another reason for choosing ATIS was that it consists of
several thousand sentences in a constrained discourse
domain, which helped avoid sparseness problems in training
collocation functions.\footnote{The hand-parsed sub-corpus was that
on the ACL DCI CD-ROM 1
of September 1991. The larger corpus, used for the bulk of the work
reported here, consisted of 4615 class A and D sentences from the
ATIS-2 training corpus. These were all such sentences of up to
fifteen words that we had access to at the time, excluding a set
of randomly selected sentences that were set aside for other
testing purposes.}

In the various experiments, the alternatives we are choosing
between are analyses expressed in the version of quasi logical
form (QLF) described by Alshawi and Crouch (1992).
QLFs express semantic content, but are derived compositionally from
complete syntactic analyses of a sentence and therefore mirror much
syntactic structure as well. However,
the use of QLF analyses is not central to our method: the important
thing is that the representation used is rich enough to support a
variety of preference functions.
We have experimented with combinations of around thirty different
functions, and use twenty of them in our spoken language translation
system; the others contribute so little to overall performance that
their computational cost cannot be justified. This default
set of twenty was used throughout the scaling factor work described
in sections \ref{scaling} and \ref{analysis}. It consists of one
collocation-based function and nineteen non-collocation-based ones.
The work described in section \ref{collcompare} involved
substituting single alternative collocation-based functions for the single
one in the set of twenty.

Many (unscaled) preference functions simply return integers
corresponding to counts of particular constructs in the representation,
such as the number of expressions corresponding to adjuncts,
unresolved ellipsis, particular attachment configurations, or
balanced conjunctions. There are also some real-valued functions,
including the semantic collocation functions discussed
in section \ref{semcoll}.

To illustrate how the system works, consider the ATIS
sentence ``Do I get dinner on this flight?''.
The CLE assigns two
analyses to this sentence; in one of them, $QH$, ``on this flight''
attaches high to ``get'', and in the other, $QL$, it attaches low to
``dinner''.
Four functions return non-zero scores on these analyses.
Two of them, \verb!Low1! and \verb!Low2!, prefer low attachment;
the difference between them is an implementation detail which can be
ignored here. A third,
\verb!SynRules!, returns an estimate of
the log probability of the syntactic rules used to construct the
analysis. A fourth, \verb!SemColl!, is a semantic collocation
function.
The scores, after multiplying by scaling factors, are as shown in
Table \ref{dinner}.
\begin{table}
\begin{center}
\begin{tabular}{|c|r|r|}\hline
Function          & Score & Score \\
                & on $QH$ & on $QL$ \\ \hline
\verb!Low1!     & -9.08 & -4.03 \\ \hline
\verb!Low2!     & -2.80 &  0.00 \\ \hline
\verb!SynRules! & -13.08 & -12.78 \\ \hline
\verb!SemColl!  & 24.32 & 3.38 \\ \hline
TOTAL           & -0.64 & -13.34 \\ \hline
\end{tabular}
\caption{Scaled preference scores for ``Do I get dinner on this
flight?''}
\label{dinner}
\end{center}
\end{table}
The \verb!SemColl! function is the only one that prefers $QH$ to $QL$.
Because this function has a relatively large scaling factor, it is able
to override the other four, which all prefer $QL$ for syntactic
reasons.

\subsection*{Training Data}

The Penn Treebank contained around 650 ATIS trees, which
we used during initial development of training and optimisation
software.
Some of the results in these initial trials were encouraging,
but most appeared to
be below reasonable thresholds of statistical significance. So, we
concluded that it was worthwhile to produce more training data.  For
this purpose we developed a semi-automatic mechanism for producing
skeletal constituent structure trees directly from QLF analyses
proposed by our analyser. In order to make these trees compatible
with the treebank and also to make them relatively insensitive to
minor changes in semantic analysis, these QLF-induced trees simply
consist of nested constituents with two categories, \verb!A! (argument) and
\verb!P! (predication), corresponding to constituents induced by QLF
\verb!term! and \verb!form! expressions respectively.
The tree for the example sentence used above is:
\begin{quote}\begin{verbatim}
(P do
   (A I)
   get
   (A dinner)
   (P on
       (A this
          flight)))
\end{verbatim}\end{quote}

The interactive software for producing the trees proposes constituents
for confirmation by a user, and takes account of answers given, to
minimize the number of interactive choices that need to be made.
Of the 4615 sentences in our training set, the CLE produced an
acceptable constituent structure for 4092 (about 89\%). A skeletal
tree for each of these 4092 sentences was created in this way and used in
the various experiments whose results are described below. We do not
directly address here the problems of applying preference functions to
select the best analysis when none is completely correct; we assume,
based on our experience with the spoken language translator, that
functions and scaling factors trained on cases where a completely
correct analysis exists will also perform fairly well on cases where
one does not.

\subsection*{Training Score}

Employing treebank analyses in the training process required
defining a measure of the ``degree of correctness'' of a QLF analysis
under the assumption that the phrase-structure analysis in the
treebank is correct.  At first sight this might appear difficult, in
that QLF is a logical formalism, but in fact it preserves much of the
geometry of constituent structure.  Specifically, significant
(typically BAR-2 level) constituents tend to give rise to \verb!term!
(roughly argument) or \verb!form!  (roughly predication) QLF
subexpressions, though the details do not matter here. It is thus
possible to associate segments of the input with such QLF
subexpressions and to check whether such a segment is also present as
a constituent in the treebank analysis. The issues raised by measuring
QLF correctness in terms of agreement with structures containing less
information than those QLFs are discussed further at the end of
section \ref{analysis}.

The training score functions we considered for a QLF $q$ with respect
to a treebank tree $t$ were functions of the form
\begin{quote}
$score(q,t) = a_1 \mid Q \cap T \mid -
                 a_2 \mid Q  \setminus  T \mid -
                 a_3 \mid T  \setminus   Q \mid $
\end{quote}
where $Q$ is the set of string segments induced by the \verb!term!
and \verb!form! expressions of $q$, $T$ is the set of constituents
in $t$, $a_1$, $a_2$, and $a_3$ are positive constants, and the
``$\setminus$'' operator denotes set difference. The
idea is to reward the QLF for constituents in common with the treebank
to penalize it for differences. Trial and error led us to
choose
\begin{quote}
$a1=1$, $a2=10$, $a3=0$
\end{quote}
which penalizes hallucination of incorrect constituents (modeled by $
\mid Q \setminus T \mid $) more heavily than a shortfall in
completeness (modeled by $ \mid Q \cap T \mid $). These constants were
fixed before we carried out the experiments whose results are
presented below.

The explanation for setting $a_3$ to $0$ was that trees
in the Penn Treebank contain many constituents that
do not correspond to QLF \verb!form! or \verb!term! expressions;
we had to avoid penalizing QLF analyses simply because
the treebank uses a different kind of linguistic representation.
For QLF-induced trees, where the correspondence is one-to-one,
it is also reasonable to set $a_3$ to 0 because when $ \mid T
\setminus Q \mid $
is non-zero, $ \mid Q \cap T \mid $ tends to be non-maximal. Among the
4092 sentences for which skeletal trees were derived, there were only
five with alternative QLFs for which the training score value was
the same with $a_3=0$ but would be different if $a_3$ were non-zero.

\section{Computing Scaling Factors}
\label{scaling}

When we first implemented a disambiguation mechanism of the kind described
above, an initial set of scaling factors was chosen by hand according to
knowledge of how the particular raw preference functions were
computed and introspection about the `strength' of the functions as
indicators of preference. These initial scaling
factors were subsequently revised according to their observed
behaviour in ranking analyses, eventually leading to reasonably well behaved
rankings.

However, as suggested earlier, there are a number of disadvantages to
manual tuning of scaling factors. These include the effort spent in
maintaining the parameters; this effort is greater for those with
less knowledge of how the raw preference functions are computed, since
this increases the effort for trial-and-error tuning. A point of
diminishing returns is also reached after which further attempts at
improvement through hand-tuning often turn out to be
counter-productive. Another problem was that it became difficult to
detect preference functions that were ineffective, or simply wrong, if
they were given sufficiently low scaling factors.  Probably a more
serious problem is that the contributions of different preference
functions to selecting the most plausible analyses seem to vary from one
sublanguage to another.  These disadvantages point to the need for
automatic procedures to determine scaling factors that optimise
preference function rankings for a particular sublanguage.

In our framework, a numerical `preference score' is computed for each of the
alternative analyses, and the analyses are ranked according to this
score. As mentioned earlier, the preference score is a weighted sum
of a set of preference functions:
Each preference function $f_j$ takes a complete QLF representation $q_i$ as
input,
returning a numerical score $s_{ij}$, the overall preference score being
computed by summing over the
product of function scores with their associated scaling factors $c_j$:
\begin{quote}
$c_1 s_{i1} + \ldots + c_m s_{im}$
\end{quote}

\subsection*{Collection Procedure}

The training process begins by analysing the corpus
sentences and computing, for each analysis of each sentence, the
training score of the analysis with respect to the manually-approved
skeletal tree and the (unscaled) values of the preference functions
applied to that analysis.

One source of variation in the data that we want to ignore in order to
derive scaling factors appropriate for selecting QLFs is the fact that
preference function values for an analysis often reflect
characteristics shared by all analyses of a sentence, as much as the
differences between alternative analyses. For example, a function that
counts the occurrences of certain constructs in a QLF will tend to
give higher values for QLFs for longer sentences. In the limit, one
can imagine a function $\phi$ that, for an $N$-word sentence, returned a value
of $N+G$ for a QLF with training score $G$ with respect to the skeletal
tree.  Such a function, if it existed, would be extremely useful, but
(if sentence length were not also considered) would not be a
particularly accurate predictor of QLF training score.

In order to discount irrelevant intersentence variability, both the
training score with respect to the skeletal tree and all the preference
function scores are therefore {\it relativised} by subtracting from
them the corresponding values for the analysis of that sentence which
best matches the skeletal tree.  If the best match is shared by
several analyses, the average for those analyses is subtracted. The
relativised training score is the distance function with respect to
which the first stage of scaling factor calculation takes place.  It
can be seen that the relativised results of our hypothetical
preference function $\phi$ are a perfect predictor of relativised
training score. Consider, for example, a six-word sentence with three QLFs,
two of which, $q_1$ and $q_2$, have completely correct skeletal tree
structures, and the third of which, $q_3$, does not. Suppose also that
the training scores and the scores assigned by preference functions,
$\phi$, $f_1$ and $f_2$ are as follows:
\begin{center}
\begin{tabular}{|c|c|c|c|c|}\hline
     & Training & $\phi$ & $f_1$ & $f_2$ \\ \hline
$q_1$ &   10     &   16   &   8   &  4   \\ \hline
$q_2$ &   10     &   16   &   6   &  10  \\ \hline
$q_3$ &    4     &   10   &   2   &  12  \\ \hline
\end{tabular}
\end{center}
After relativizing (subtracting the average of the $q_1$ and $q_2$
values) we get
\begin{center}
\begin{tabular}{|c|c|c|c|c|}\hline
     & Training & $\phi$ & $f_1$ & $f_2$ \\ \hline
$q_1$ &    0     &    0   &   1   &  -3  \\ \hline
$q_2$ &    0     &    0   &  -1   &   3 \\ \hline
$q_3$ &   -6     &   -6   &  -5   &   5 \\ \hline
\end{tabular}
\end{center}

\subsection*{Least Squares Calculation}
\label{leastsquares}

An initial set of scaling factors is calculated in a straightforward
analytic way by approximating $g_i$, the relativised training score of
$q_i$, by $\sum_j c_j z_{ij}$, where $c_j$ is the scaling factor for
preference function $f_j$ and $z_{ij}$ is the relativised score
assigned to $q_i$ by $f_j$. We vary the values of $c_j$ to minimize
the sum, over all QLFs for all training sentences, of the squares of
the errors in the approximation,
\begin{quote}\begin{math}
\sum_i (g_i - \sum_j  c_j  z_{ij}  )^2
\end{math}\end{quote}
Defining the error function as a sum of squares of differences in this
way means that the minimum error is attained when the derivative with
respect to each $c_k$, $-2 \sum_i z_{ik} ( g_i - \sum_j c_j z_{ij} )$,
is zero.  These linear simultaneous equations, one for each of $c_1
\ldots c_m$, can be solved by Gaussian elimination. (For a full
explanation of this standard technique, see Moore and McCabe, 1989,
pp. 174ff and 680ff.)

This least-squares set of scaling factors achieves quite good
disambiguation performance (see section \ref{analysis} below) but is
not truly optimal because of the inherent nonlinearity of the goal,
which is to maximize the proportion of sentences for which a correct
QLF is selected, rather than to approximate training scores (even
relativised ones).  Suppose that a function $M$ has a tendency to give
high scores to correct QLFs when the contributions of other functions do
not clearly favour any QLF, but that $M$ tends to perform much less
well when other functions come up with a clear choice. Then increasing
the scaling factor on $M$ from the least-squares value will tend to
improve system performance even though the sum of squares of errors is
increased; $M$'s tendency to perform well just when it is important to
do so should be rewarded.

\subsection*{Iterative Scaling Factor Adjustment}
\label{hillclimbing}

The least-squares scaling factors are therefore adjusted iteratively
by a hill-climbing procedure that directly examines the QLF choices
they give rise to on the training corpus. Scaling factors are altered
one at a time in an attempt to locally optimise\footnote{Finding a global
optimum would of course be desirable. However, inspection of the
results, over various conditions, of the iterative scheme presented
here did not suggest that the introduction of a technique such as
simulated annealing, which in general can improve the prospect of
finding a more global optimum, would have had much effect on performance.}
the number of correct disambiguation decisions, i.e.\ the number of training
sentences for which a QLF with a correct skeletal tree receives the
highest score.

A step in the iteration involves calculating the effect of an
alteration to each factor in turn.\footnote{An algorithm based on
gradient descent might appear preferable, on the grounds that it would
alter all factors simultaneously and have a better chance of locating
a global optimum.  However, the objective function, the number of
correct disambiguation decisions, varies discontinuously with the
scaling factors, so no gradients can be calculated.} If factors $c_k,
k \ne j$ are held constant, it is easy to find a set (possibly empty)
of real-valued intervals $[u_{ij},v_{ij}]$ such that a correct choice
will be made on sentence $i$ if $u_{ij} \leq c_j \leq v_{ij}$. By
collecting these intervals for all the functions and for all the
sentences in the training corpus, one can determine the effect on the
number of correct disambiguation decisions of any alteration to any
single scaling factor. The alteration selected is the one that gives
the biggest increase in the number of sentences for which a correct
choice is made. When no increase is possible, the procedure
terminates. We found that convergence tends to be fairly rapid, with the number
of
steps seldom exceeding the number of scaling factors involved
(although the process does occasionally change a scaling factor it has
previously altered, when intervening changes make this
appropriate).

One of the functions we used shows the limitations of least-squares scaling
factor optimisation, alluded to above, in quite a dramatic way. The
function in question returns the number of temporal modifiers in a QLF.
Its intended purpose is to favour readings of utterances like
``Atlanta to Boston Tuesday'' where ``Tuesday'' is a temporal modifier
of the (elliptical) sentence rather than forming a compound noun with
``Boston''. Linear scaling always gives this function a negative weight,
causing temporal modifications to be downgraded, and in fact
the relativised training score of a QLF turns out to be negatively
correlated with the number of temporal modifiers it contains. However,
the intuitions that led to the introduction of the function do seem to
hold for QLFs that are close to being correct, and therefore iterative
adjustment makes the weight positive.

\section{Comparing Scaling Factor Sets}
\label{analysis}

The performance of the factors derived from least squares calculation
and adjustment by hill-climbing was compared with that of various
other sets of factors. The factor sets considered, roughly
in increasing order of their expected quality, were:
\begin{itemize}
\item ``Normalized'' factors: the magnitude of each factor is
the inverse of the standard deviation of the preference function
in question, making each function contribute equally. A factor is
positive if it correlates positively with training scores, negative
otherwise.
\item Factors chosen and tuned by hand for ATIS sentences
before the work described in this paper was done, or, for functions
developed during the work described here, without reference to any
automatically-derived values.
\item Factors resulting from least squares calculation, as described
in section \ref{leastsquares} above.
\item Factors resulting from least squares calculation followed by
hill-climbing adjustment (section \ref{hillclimbing} above).
\end{itemize}
To provide a baseline, performance was also evaluated for the
technique of a random selection of a single QLF for each sentence.

The performance of each set of factors was evaluated as follows.  The
set of 4092 sentences with skeletal trees was divided into five
subsets of roughly equal size. Each subset was ``held out'' in turn:
the functions and scaling factors were trained on the other four
subsets, and the system was then evaluated on the held-out subset.
The system was deemed to have correctly processed a sentence if the
QLF to which it assigned the highest score agreed exactly with the
corresponding skeletal tree.

The numbers of correctly-processed sentences (i.e.\ sentences whose
selected QLFs had correct constituent structures) are shown in Table
\ref{scores1}; because all the sentences involved were within
coverage, the theoretical maximum achievable is 4092 (100\%).

\begin{table}
\begin{center}
\begin{tabular}{|l|c|c|} \hline
Scaling factor set & Number  & Percentage \\
                   & correct & correct \\ \hline
(Random baseline)  &  1949   & 47.6   \\ \hline
Normalized         &  3549   & 86.7   \\ \hline
Hand tuned         &  3717   & 90.8   \\ \hline
Least squares      &  3841   & 93.9   \\ \hline
Hill-climbing      &  3857   & 94.3   \\ \hline
\end{tabular}
\caption{Performance of scaling factor sets}
\label{scores1}
\end{center}
\end{table}
We use a standard statistical method, the `sign test'
(explained in, for example, Dixon and Massey 1968) to assess the
significance of the difference between two factor sets $S_1$
and $S_2$. Define $F_i(x)$ to be the
function that assigns 1 to a sentence $x$ if $S_i$ makes the
correct choice in disambiguating $x$, and 0 if it makes the wrong
choice.
The null hypothesis is that $F_1(x)$ and $F_2(x)$, treated
as random variables over $x$, have the same distribution,
from which we would expect the difference between
$F_1(x)$ and $F_2(x)$ to be positive as often as it is negative.
Table \ref{signtest} gives the number of cases
in which this difference is positive or negative. As is usual for
the sign test, the cases where the difference is 0 do not
need to be taken into account. The test is applied to compare
six pairs of factor sets. The ``\#SDs'' column shows the number
of standard deviations represented by the difference between the
``+'' and ``--'' figures under the null hypothesis; a \#SDs value
of 1.95 is statistically significant at the 5\% level (two-tail),
and a value of 3.3 is significant at the 0.1\% level.

\begin{table}
\begin{center}
\begin{tabular}{|l|l|c|c|c|} \hline
$S1$    &  $S2$                & $+$ & $-$ & \#SDs \\ \hline
Normalized    & Hand-tuned     & 154 & 322 & 7.7 \\
Normalized    & Least squares  & 67  & 359 & 14.1 \\
Normalized    & Hill climbing  & 75  & 383 & 14.4 \\
Hand-tuned    & Least squares  & 78  & 202 & 7.4 \\
Hand-tuned    & Hill climbing  & 76  & 216 & 8.2 \\
Least squares & Hill climbing  & 20  &  36 & 2.1 \\ \hline
\end{tabular}
\caption{Sign test comparisons of scaling factor sets}
\label{signtest}
\end{center}
\end{table}

Table \ref{signtest} shows that, in terms of wrong QLF choices, both sets of
machine-optimised factors perform significantly better than the
hand-optimised factors, to which considerable skilled effort had been
devoted. It is worth emphasising that the process of determining the
machine-optimised factors does not make use of the knowledge encoded
by hand optimisation. The hill climbing factor set, in turn,
performs significantly better than the least squares set from which
it is derived.

A possible objection to this analysis is that, because QLFs are much
richer structures than constituent trees, it is possible for a QLF to
match a tree perfectly but have some other characteristic that makes
it incorrect. In general, the principle source of such discrepancies is a
wrong choice of word sense, but pure sense ambiguity (i.e.\ different
predicates for the same syntactic behaviour of the same word) turns
out to be extremely rare in the ATIS corpus. An examination of the
selected QLFs for the 20+36=56 sentences making up the $+$ and $-$
values for the comparison between the least squares and hill climbing
factor sets showed that in {\it no} case did a QLF have a correct
constituent structure but fail to be acceptable on other criteria.
Thus while the absolute percentage correctness figures for a set of
scaling factors may be very slightly (perhaps up to 1\%)
overoptimistic, this has no noticeable effect on the {\it differences}
between factor sets.

\section{Lexical Semantic Collocations}
\label{semcoll}

In this section we move from the problem of calculating scaling
factors to the other main topic of this paper,
showing how our experimental framework can be used
diagnostically to compare the utility of competing suggestions for
preference functions. We refer
to the variant of collocations we used as lexical semantic collocations
because (i) they are collocations between word senses rather than
lexical items, and (ii) the relationships used are often deeper
than syntactic relations (for example the relations between a
verb and its subject are different for passive and active sentences).

The semantic collocations extracted from QLF expressions take
the form of $(H1,R,H2)$ triples where $H1$ and $H2$ are
the head predicates of phrases in a
sentence and $R$ indicates the relationship (e.g. a
preposition or an argument position) between the two phrases in the
proposed analysis. For this purpose, the triple derivation software
abstracted away from proper names and some noun and verb predicates
when they appeared as heads of
phrases, replacing them by hand-coded class predicates.
For example, predicates for names of meals are mapped onto the class name
\verb!cc_SpecificMeal! on the grounds that their distributions
in unseen sentences are likely to be very similar.

Some of the
triples for the high-attachment QLF for ``Do I get dinner on this
flight?'' are:
\begin{quote}\begin{verbatim}
(get_Acquire,2,personal)
(get_Acquire,3,cc_SpecificMeal)
(get_Acquire,on,flight_AirplaneTrip)
\end{verbatim}\end{quote}
The first two triples correspond to the agent and theme positions in
the predicate for \verb!get!, while the third expresses the vital PP
attachment. In the triple set for the other QLF, this triple is
replaced by
\begin{quote}\begin{verbatim}
(cc_SpecificMeal,on,flight_AirplaneTrip).
\end{verbatim}\end{quote}

Data collection for the semantic collocation
functions proceeds by deriving a set of triples from each QLF analysis
of the sentences in the training set. This is followed by statistical
analysis to produce the following functions of each triple in the
observed triple population. The first two functions have been used in
other work on collocation; some authors use simple pairs rather than
triples (i.e.\ no relation, just two words) when computing collocation
strengths, so direct comparisons are a little difficult.
The third function is an original variant of the
second; the fourth is original; and the fifth is prompted by the
arguments of Dunning (1993).
\begin{itemize}

\item Mutual information: this relates the probability
$P_1(a) P_2(b) P_3(c)$ of the triple $(a,b,c)$ assuming independence
between its three fields, where $P_p(x)$ is the probability of
observing $x$ in position $p$, with the probability $A$
estimated from actual observations of triples derived from analyses
ranked highest (or joint highest) in training score.
More specifically, we use $ln(A/(P_1(a)P_2(b)P_3(c)))$.

\item $\chi^2$: compares the expected frequency $E$ of a
triple with the square of the difference between $E$ and the observed
frequency $F$ of the triple.
Here the observed frequency is in analyses ranked highest
(or joint highest) in training score and the
``expected'' frequency assumes independence between triple fields.
More specifically we use $\frac{|F-E|*(F-E)}{E}$.
This variant of $\chi^2$, in which the numerator is
signed, is used so that the function is monotonic, making it
more suitable in preference functions.

\item $\chi$: as $\chi^2$, but the quantity used is
$\frac{(F-E)}{\sqrt{E}}$, as large values of $F-E$ have a tendency
to swamp the $\chi^2$ function.

\item Mean distance: the average of the relativised training score
for all QLF analyses (not necessarily highest ranked ones) which
include the semantic collocation corresponding to the triple.  In
other words, the mean distance value for a triple is the mean amount
by which a QLF giving rise to that triple falls short of a perfect
score.

\item Likelihood ratio: for each triple $(H1,R,H2)$, the ratio of the
maximum likelihood of the triple, given the distribution of triples in correct
analyses of the training data, on the assumption
that $H1$ and $H2$ are
independent given $R$, to the maximum likelihood without that
assumption. (See Dunning, 1993, for a fuller explanation of the
use of likelihood ratios).

\end{itemize}

Computation of the mutual information and $\chi^2$ functions
for triples involves the simple smoothing technique, suggested
by Ken Church, of adding 0.5 to actual counts.

 From these five functions on triples we define five semantic
collocation preference functions applied to QLFs, in each case by
averaging over the result of applying the function to each triple
derived from a QLF. We refer to these functions by the same names as
their underlying functions on triples. The collocation functions
are normalized by multiplying up by the number of words in the
sentence to which the function is being applied. This normalization
keeps scores for QLFs in the same sentence comparable, while
at the same time ensuring the triple function scores tend to grow
with sentence length in the same way that the non-collocation
functions tend to do. Thus the optimality of a set of scaling
factors is relatively insensitive to sentence length.

Our use of the mean distance function was motivated by the desire
to take into account additional information from the training
material which is not exploited by the other collocation functions.
Specifically, it takes into account {\it all} analyses proposed
by the system, as well as the magnitude of the training score.
In contrast, the other collocation
functions only make use of the training score to select the best
analysis of a sentence, discarding the rest. Another way of putting
this is that the mean distance function is making use of negative
examples and a measure of the degree of unacceptability of an
analysis.

\section{Comparing Semantic Collocation Functions}
\label{collcompare}

An evaluation of each function acting alone on the five held-out sets of
test data yielded the numbers of correctly-processed sentences shown
in Table \ref{semcollalone}.
The figures for the random baseline are repeated from
Table \ref{scores1}. We also show, for comparison, the results for a
function that scores a QLF according to the sum of the logs of the
estimated probabilities of the syntactic rules used in its
construction.\footnote{We estimate the probability of occurrence of
a syntactic rule $R$ as the number of occurrences of $R$ leading to
QLFs with correct skeletal trees, divided by the number of
occurrences of all rules leading to such QLFs.}
\begin{table}
\begin{center}
\begin{tabular}{|l|c|c|} \hline
Function             & Number  & Percentage \\
                    & correct & correct \\ \hline
(Random baseline)   & 1949    & 47.6   \\ \hline
Mutual info         & 2817    & 68.9   \\ \hline
Syntactic rule cost & 2913    & 71.2   \\ \hline
L'hood ratio        & 3120    & 76.3   \\ \hline
$\chi^2$            & 3339    & 81.6   \\ \hline
$\chi$              & 3407    & 83.3   \\ \hline
Mean distance       & 3670    & 89.7   \\ \hline
\end{tabular}
\caption{Performance of collocational and syntactic rule functions alone}
\label{semcollalone}
\end{center}
\end{table}

To arrive at the figures shown, where a function
judged $N$ QLFs equally plausible, of which $0 < G < N$ were correct,
we assigned a fractional count $\frac{G}{N}$ to that sentence;
a random choice among the $N$ QLFs would pick a
correct one with probability $\frac{G}{N}$. For significance tests,
which require binary data, we took a function as performing correctly
only if {\it all} the QLFs it selected were correct.  Such ties
did not occur at all for the other experiments reported in this paper.

A pairwise comparison of the results shows that all the differences
between collocational functions are statistically highly significant.
The syntactic rule cost function is significantly worse than all the
collocational functions except the mutual information one, for which the
difference is not significant either way. (There may, of course, exist
better syntactic functions than the one we have tried.)
The mean distance function is
much superior to all the others when acting alone.
Presumably, this function has an edge over the other
functions because it exploits the additional information
from negative examples and degree of correctness.

The difference in performance between our syntactic and semantic
preference functions is broadly in line with the results
presented by Chang, Luo, and Su (1992) who use probabilities
of semantic category tuples. However, this similarity in the
results should be taken with some caution, because our syntactic
preference function is rather crude, and because our best semantic
function (mean distance) uses the additional information
mentioned above. This information is not normally taken into account by
direct estimates of tuple probabilities.

When one collocation function is selected to act together with the
nineteen non-collocation-based functions from the default set (the set
defined in section \ref{setup} and used in the experiments on scaling
factor calculation) the picture changes slightly.  In this context,
when scaling factors are calculated in the usual way, by least-squares
followed by hill-climbing, the results for the best three of the above
functions are as shown in Table \ref{semcolltogether}.

\begin{table}
\begin{center}
\begin{tabular}{|l|c|c|} \hline
Function             & Number  & Percentage \\
                   & correct & correct \\ \hline
$\chi^2$           &  3741   & 91.4   \\ \hline
$\chi$             &  3766   & 92.0   \\ \hline
Mean distance      &  3857   & 94.3   \\ \hline
\end{tabular}
\caption{Performance of collocational functions with others}
\label{semcolltogether}
\end{center}
\end{table}

The difference between the mean distance function and the other two is
still highly significant; therefore this function is chosen to be
the only collocational one to be included in the default set of twenty
(hence the ``mean distance'' condition here is the same as the
``hill-climbing'' condition in section \ref{analysis}).  However, the
difference between the $\chi$ and $\chi^2$ functions is no longer quite
so clear cut, and the relative advantage of the mean distance function
compared with the $\chi$ function is less. It may be that other preference
functions make up for some shortfall of the $\chi$ function that is, at least
in part, taken account of by the mean distance function.

\section{Conclusion}

We have presented a relatively simple analytic technique for
automatically determining a set of scaling factors for preference
functions used in semantic disambiguation. The initial scaling factors
produced are optimal with respect to a score provided by a
training procedure, and are further improved by comparison with
instances of the task they are intended to perform.  The experimental
results presented indicate that, by using a fairly crude training score
measure (comparing only phrase structure trees) with a few thousand
training sentences, the method can yield a set of scaling factors that
are
significantly better than
those derived by a labour intensive hand tuning effort.

We have also confirmed empirically that considerable differences exist
between the effectiveness of differently formulated collocation
functions for disambiguation. The experiments provide a basis for
selecting among different collocational functions, and suggest that a
collocation function must be evaluated in the context of other functions,
rather than on its own, if the correct selection is to be made.

It should be possible to extend this work fruitfully in several
directions, including the following.
Training with a measure defined directly on semantic
representations is likely to lead to a further reduction in
the disambiguation error rate.
The method for computing scaling factors described here has more
recently been applied to optimising preference selection for the task of
choosing between analyses arising from different word hypotheses in
a speech recognition system (Rayner {\it et al}, 1994),
and is applicable to other problems, such as choosing between
possible target representations in a machine translation system.
Finally it would be interesting to combine the work on semantic
collocation functions with that on similarity-based clustering
(Pereira, Tishby and Lee 1993; Dagan, Marcus and Markovitch 1993)
with the aim of overcoming
the problem of sparse training data. If this is successful, it
might make these functions suitable for disambiguation in domains
with larger vocabularies than ATIS.

\section*{Acknowledgements}

We would like to thank Manny Rayner for many useful suggestions in
carrying out this work, for making the selections necessary to create
the database of skeletal trees, and for helping with inspection of
experimental results. The paper has also benefited from discussions
with Fernando Pereira, Ido Dagan, Michael Collins, Steve Pulman and
Jaan Kaja, and from the comments of four anonymous CL referees.

Most of the work reported here was carried out under a Spoken Language
Translation project funded by Telia (formerly Televerket/Swedish
Telecom).  Other parts were done under the CLARE project (JFIT project
IED4/1/1165), funded by the UK Department of Trade and Industry, SRI
International, the Defence Research Agency, British Telecom, British
Petroleum and British Aerospace.

\section*{References}

\begin{reverseindent}

\item\pagebreak[3]
Agn\"{a}s, M-S., {\it et al}. 1993. {\it Spoken Language Translator:
First Year Report}. SRI International Cambridge Computer Science
Research Centre, Report 043.

\item\pagebreak[3]
Alshawi, H.  and R.  Crouch.  1992.  ``Monotonic Semantic Interpretation''.
{\it Proceedings of the 30th Annual Meeting of the Association for
Computational Linguistics}, Newark, Delaware, 32--39.

\item\pagebreak[3]
Alshawi, H., ed. 1992. {\it The Core Language Engine}. Cambridge,
Massachusetts: The MIT Press.

\item Calzolari, N. and R. Bindi. 1990. ``Acquisition of Lexical Information
from a Large Textual Italian Corpus''. {\it Proceedings of the 13th
International Conference on Computational Linguistics}, 3:54--59.

\item\pagebreak[3]
Chang, J., Y. Luo, and K. Su. 1992. ``GPSM: A Generalized Probabilistic
Semantic Model for Ambiguity Resolution''. {\it Proceedings of the 30th
Annual Meeting of the Association for Computational Linguistics}, 177--192.

\item\pagebreak[3]
Church, K.W. and P. Hanks. 1990. ``Word Association Norms, Mutual Information,
and Lexicography''. {\it Computational Linguistics} 16:22--30.

\item\pagebreak[3]
Dagan, I., S. Marcus and S. Markovitch. 1993. ``Contextual Word
Similarity and Estimation from Sparse Data''.
{\it Proceedings of the 31st meeting of the
Association for Computational Linguistics}, ACL, 164--171.

\item\pagebreak[3]
Dixon, W.J. and F.J. Massey. 1968. {\it Introduction to Statistical
Analysis}, third edition. New York: McGraw-Hill.

\item\pagebreak[3]
Dunning, T. ``Accurate Methods for Statistics
of Surprise and Coincidence''. {\it Computational Linguistics} 19:61--74.

\item\pagebreak[3]
Gale, W.A. and K.W. Church. 1991. ``Identifying Word Correspondences in
Parallel Texts''. {\it Proceedings of the DARPA Speech and Natural Language
Workshop}, Morgan Kaufmann, 152--157.

\item\pagebreak[3]
Hindle, D. and M. Rooth. 1993. ``Structural Ambiguity and Lexical
Relations''. {\it Computational Linguistics}. 19:103--120.

\item\pagebreak[3]
Hobbs, J.~R. and J.~Bear. 1990. ``Two Principles of Parse Preference''. Vol. 3,
{\it Proceedings of the 13th International Conference on Computational
Linguistics}, Helsinki, 162--167.

\item\pagebreak[3]
Marcus, M.P., B. Santorini, and M.A. Marcinkiewicz. 1993. ``Building a Large
Annotated Corpus of English: the Penn Treebank''.
{\it Computational Linguistics} 19:313--330.

\item\pagebreak[3]
McCord, M.C. 1989. ``A New Version of Slot Grammar''. IBM Research
Report RC 14506, IBM Thomas J. Watson Research Center, Yorktown Heights,
New York.

\item\pagebreak[3]
McCord, M.C. 1993. ``Heuristics for Broad-Coverage Natural Language
Parsing''. {\it Proceedings of the ARPA Human Language Technology
Workshop}. Los Altos: Morgan Kaufmann, 127--132.

\item\pagebreak[3]
Moore, D.~S. and G.~P. McCabe. 1989. {\it Introduction to the Practice of
Statistics}. New York/Oxford: Freeman.

\item\pagebreak[3]
Ostendorf, M., A. Kannan, S. Austin, O. Kimball, R. Schwartz,
J.R. Rohlicek. 1991. ``Integration of Diverse Recognition Methodologies
Through Reevaluation of N-Best Sentence Hypotheses''.
{\it Proceedings of the DARPA Speech and Natural Language
Workshop}, 83--87.

\item\pagebreak[3]
Pereira, F., N. Tishby and L. Lee. 1993. ``Distributional Clustering
of English Words''. {\it Proceedings of the 31st meeting of the
Association for Computational Linguistics}, ACL, 183--190.

\item\pagebreak[3]
Rayner, M., D.~M. Carter, V. Digalakis and P. Price. 1994 (to appear).
``Combining Knowledge Sources to Reorder N-best Speech Hypothesis Lists'',
{\it Proceedings of the ARPA HLT Meeting}.

\item\pagebreak[3]
Resnik, P. and M.~A. Hearst. 1993. ``Structural Ambiguity and
Conceptual Relations''. {\it Proceedings of the Workshop on
Very Large Corpora}, ACL, 58--64.

\item\pagebreak[3]
Sekine, S., J.~J. Carroll, S. Ananiadou, and J. Tsujii. 1992. ``Automatic
Learning for Semantic Collocation''. {\it Proceedings of the Third Conference
on Applied Natural Language Processing}, ACL, 104--110.

\end{reverseindent}

\end{document}